\documentclass[runningheads]{llncs}
\usepackage[T1]{fontenc}
\usepackage{cite}
\usepackage{amsmath,amssymb,amsfonts}
\usepackage{algorithmic}
\usepackage{graphicx}
\usepackage{textcomp}
\usepackage{xcolor}
\usepackage{bm}
\usepackage{booktabs}
\usepackage{pifont}
\usepackage{hyperref}
\usepackage{marvosym}
\usepackage{color}

\begin{document}
\newcommand{\yes}{\ding{51}}
\newcommand{\R}{\mathbb{R}} % Real numbers
\newcommand{\rbr}[1]{\left(#1\right)}
\newcommand{\sbr}[1]{\left[#1\right]}
\newcommand{\cbr}[1]{\left\{#1\right\}}
\title{MVGT: A Multi-view Graph Transformer Based on Spatial Relations for EEG Emotion Recognition}

\titlerunning{A Multi-view Graph Transformer for EEG Emotion Recognition}
\author{Yan-Jie Cui\and
Xiao-Hong Liu\textsuperscript{(\Letter)} \and
Jing Liang \and
Ya-Min Fu}
\authorrunning{Y.-J. Cui et al.}
\institute{School of Computer Science (National Pilot Software Engineering School), \\Beijing University of Posts and Telecommunications, Beijing 100876, China
\email{\{yanjiecui,xiaohongliu,liangjing18,fuyamin\}@bupt.edu.cn}}
\maketitle              % typeset the header of the contribution
\begin{abstract}
Electroencephalography (EEG), a technique that records electrical activity from the scalp using electrodes, plays a vital role in affective computing. However, fully utilizing the multi-domain characteristics of EEG signals remains a significant challenge. Traditional single-perspective analyses often fail to capture the complex interplay of temporal, frequency, and spatial dimensions in EEG data.
To address this, we introduce a multi-view graph transformer (MVGT) based on spatial relations that integrates information across three domains: temporal dynamics from continuous series, frequency features extracted from frequency bands, and inter-channel relationships captured through several spatial encodings. This comprehensive approach allows model to capture the nuanced properties inherent in EEG signals, enhancing its flexibility and representational power.
Evaluation on publicly available datasets demonstrates that MVGT surpasses state-of-the-art methods in performance. The results highlight its ability to extract multi-domain information and effectively model inter-channel relationships, showcasing its potential for EEG-based emotion recognition tasks.

\end{abstract}

\section{Introduction}
\label{sec:intro}
Affective computing is commonly employed for the analysis of emotional states through Human-Computer Interaction (HCI) systems, which collect multimodal data from subjects, including voice signals, self-reports, body gestures and physiological signals. 
Among these modalities, physiological signals offer distinct advantages as they are directly derived from the subjects' mental states, making it difficult for them to disguise or conceal their emotions. Scalp electroencephalography (EEG), a physiological signal usually used to analyze the cognitive functions of the human brain, is collected using noninvasive electrodes on the scalp \cite{alarcao2017emotions}. 
Nowadays, due to its high temporal resolution, portability, and affordability, this method is widely employed to study brain changes in response to emotional stimuli \cite{niemic2004studies}.

EEG signals contain rich emotional information in the temporal, frequency, and spatial domains. How to reasonably and effectively utilize information from each domain is a major challenge.
Therefore, recent studies have attempted to combine multi-domain information to capture complex EEG features \cite{MD-AGCN,EmoGT,li2022multi}. Although these works have achieved good performance, their limited use of domain-specific information has constrained model performance.

The frequency domain features of EEG commonly include power spectral density (PSD), differential entropy (DE), differential asymmetry (DASM), and others. Among these, DE has been proven to be the simplest and most effective in EEG emotion recognition tasks \cite{duan2013differential}. 
DE has the ability to distinguish EEG patterns between high-frequency and low-frequency energy \cite{zheng2015investigating}.

Given the high temporal resolution of EEG signals, many studies use temporal models to extract sequential information. Recurrent neural networks (RNNs) have significantly improved emotion recognition performance by modeling temporal dependencies \cite{R2G-STNN, BiDANN}. 
However, their serial structure poses challenges for parallelization. Convolutional neural networks (CNNs) have also been applied to extract temporal features, but their performance is restricted by limited receptive fields \cite{lggNet, MD-AGCN}. 
To overcome these difficulties, \cite{EmoGT} employs the attention mechanism to capture relationships between time points; however, this method may suffer performance degradation due to time-unaligned events at a single moment. 

The spatial domain of EEG is equally crucial for understanding emotional states, as emotions often involve distributed neural circuits rather than isolated regions \cite{mauss2010measures}. Asymmetric activity between the left and right hemispheres can reflect changes in valence and arousal \cite{schmidt2001frontal}. Message-passing-based graph neural network (MPGNN) methods can effectively capture structural relationships between EEG channels \cite{rgnn, DGCNN, lggNet, MD-AGCN}. 
However, using MPGNN may pose potential risks of over-smoothing and over-squashing, which could hinder its ability to learn complex representations \cite{ller2024attending}. Additionally, these methods often underutilize the geometric and anatomical structural relationships inherent in EEG spatial information.
\begin{figure*}[htb]
   \centering
   \includegraphics[scale=0.31]{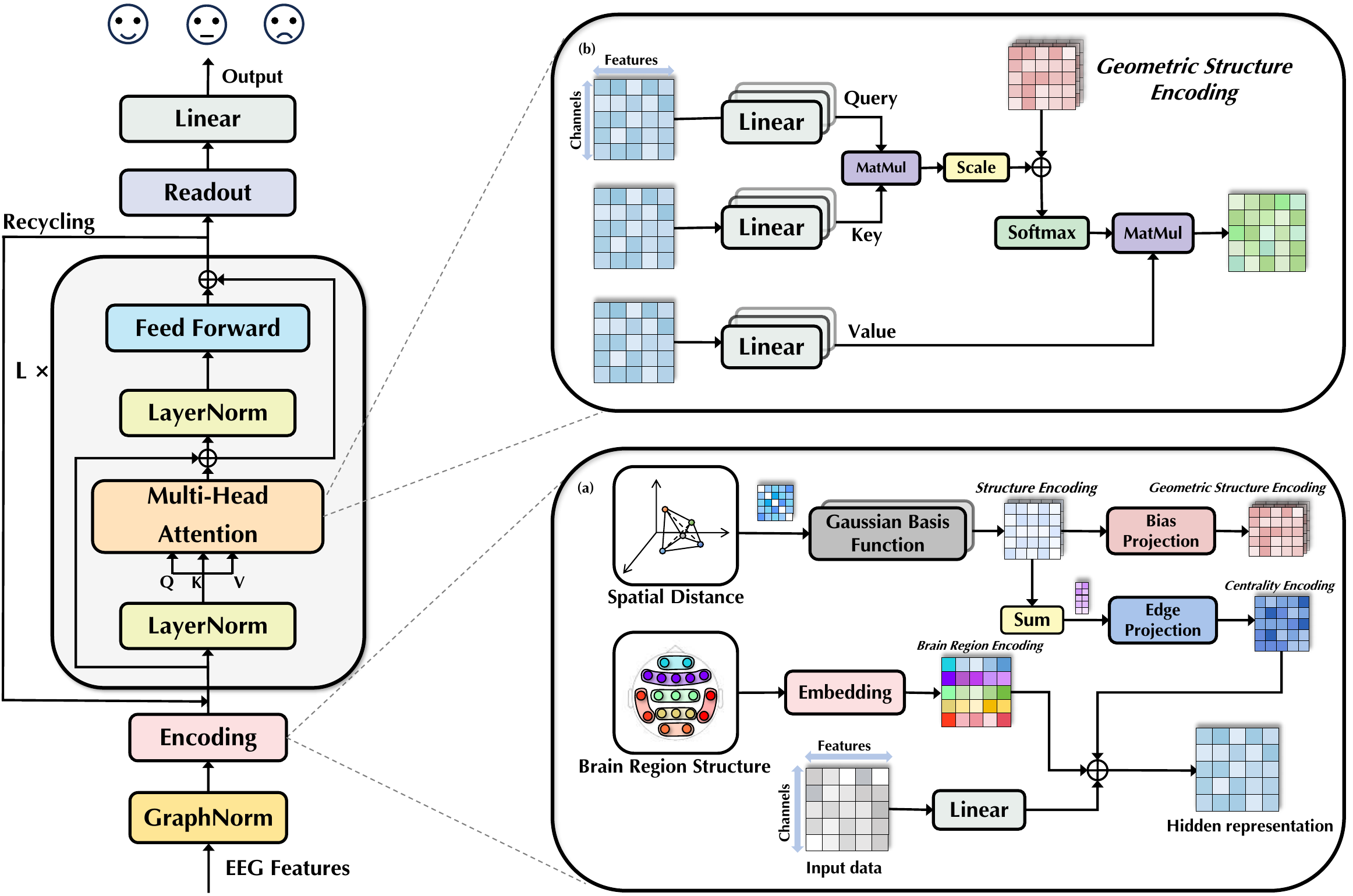}
   \caption{Overall structure of MVGT. (a) represents the process of brain region encoding, centrality encoding and geometric structure encoding. (b) depicts the process of calculating inter-channel correlations based on the attention mechanism and geometric structure encoding. ``Recycling'' refers to the iterative refinement (see \ref{sec:implementation}).}
   \label{fig:MVGT}
\end{figure*}

To address these limitations and fully exploit features in the temporal, frequency, and spatial domains, we propose a multi-view graph transformer (MVGT) based on spatial relations for EEG emotion recognition. Graph transformers offer robust and flexible spatial perception while mitigating issues like over-smoothing and over-squashing \cite{ller2024attending}. 
For the frequency domain, we use the embedding of DE as the feature representation. For the temporal domain, inspired by \cite{itransformer}, we expand the temporal receptive field by embedding entire continuous time segments as tokens and employ feed-forward neural networks to capture temporal patterns. 
In the spatial domain, we incorporate several simple yet effective spatial encodings, brain region encoding, centrality encoding and geometric structure encoding, into the model to help it capture effective spatial structures. These encodings allow MVGT to adaptively model correlations between EEG channels, delivering improved performance and deeper insights into emotional states.

\section{Methods}
\subsection{Problem Definition}
EEG signals exhibit an inherent spatial structure that can be represented as a graph $G = \rbr{V, E}$, where $V$ denotes the nodes (EEG channels), and $E$ represents the edges (connections between channels).
The features of the nodes at time $t$ are denoted by $\bm{X}_t = \rbr{\bm{x}_1, \bm{x}_2, \cdots, \bm{x}_n} \in \R^{n \times f}$, where $n = \vert V \vert$ represents the number of nodes and $f$ represents the dimension of  DE features extracted on five frequency bands, i.e., $\delta$ (1-4Hz), $\theta$ (4-8Hz), $\alpha$ (8-14Hz), $\beta$ (14-31Hz), $\gamma$ (31-50Hz). 
To model the temporal dynamics, continuous segments are extracted using a sliding window.
For the $s$-th segment, the data is represented as $\tilde{\bm{X}}_{s} = \rbr{\tilde{\bm{x}}_1, \tilde{\bm{x}}_2, \cdots, \tilde{\bm{x}}_n} \in \R^{n \times Tf}$, where $T$ is the window size.
The training samples are embedded through an encoding layer, incorporating centrality and brain region encodings to yield $\bm{H}^{0}_s = \rbr{\bm{h}_1^{0}, \bm{h}_2^{0}, \cdots, \bm{h}_n^{0}} \in \R^{n \times d}$ (see Fig.~\ref{fig:MVGT}(a)), where $d$ is the embedding dimension. The encoded data is then input into the model for emotion classification.
\begin{figure*}[htb]
   \centering
   \includegraphics[scale=0.15]{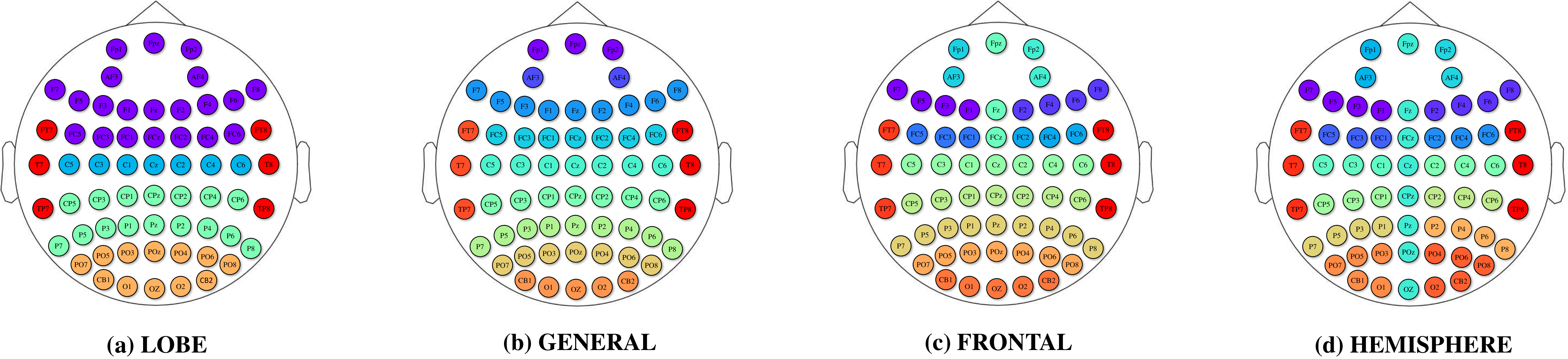}   
   \caption{The brain region division schemes are illustrated. (a) LOBE scheme shows a coarse partitioning based on lobe structures. (b) GENERAL scheme represents a fine-grained partitioning of the brain lobes. (c) FRONTAL scheme introduces symmetry of the left and right frontal regions. (d) HEMISPHERE scheme further enhances the channel symmetry in the partitioning scheme. Channels of the same color belong to the same brain region.}
   \label{fig:region}
\end{figure*}

\subsection{Temporal Embedding}
\label{sec:time}
Temporal information is critical in time-series processing. \cite{EmoGT} treated the features of different channels at the same time points as tokens and employed the attention mechanism to extract temporal correlations between them. However, due to anisotropic volume conduction characteristics \cite{craik2019deep} in human brain tissues, there may be different physical meanings between channels at the same time step, resulting time-unaligned events that subsequently degrade performance.
Other approaches \cite{MD-AGCN,lggNet} utilized CNNs to extract temporal information along the time axis from continuous EEG segments, with the receptive field limited by the size of the convolution kernel. Inspired by \cite{itransformer}, we broaden the receptive field by embedding entire time segments as tokens rather than single time points. EEG signals are segmented using overlapping sliding windows of size $T$, and these segments are treated as tokens. After processing with sliding windows, we obtain $\tilde{\bm{X}} = \rbr{\tilde{\bm{X}}_{1}, \tilde{\bm{X}}_{2}, \cdots, \tilde{\bm{X}}_{S}} \in \R^{S \times n \times Tf}$, where $S$ denotes the number of continuous EEG segments, $n$ is the number of channels, and $f$ denotes the dimension of DE features.
Refer to the Appendix \ref{appendix:time} for more temporal embedding details.

According to the universal approximation theorem \cite{hornik1991approximation}, the feed-forward neural network (FFN), as the basic module of the Transformer \cite{Transformer} encoder, can learn the intrinsic properties to describe a time series.
By processing continuous time segments as input, the FFN extracts temporal features independently for each channel, providing more advantageous predictive representations compared to self-attention \cite{itransformer}.

\subsection{Spatial Encoding}
To better recognize emotional patterns in emotion classification tasks, we employ three effective methods of spatial encoding: brain region encoding (BRE), centrality encoding (CE) and geometric structure encoding (GSE).

\subsubsection{Brain Region Encoding}
Neuroscience research indicates that specific brain regions activate together during high-level cognitive processes \cite{kober2008functional}. In EEG emotion recognition, incorporating relevant neuroscience findings can typically enhance recognition accuracy.
For example, studies like \cite{rgnn} and \cite{BiHDM} utilized the asymmetry of neural activity between the left and right hemispheres to enhance recognition performance.
\cite{lggNet} divided EEG channels into different regions and combined intra-region convolution with inter-region convolution to capture spatial relationships. 
Building on \cite{lggNet}, we propose four learnable BRE schemes based on neuroscience research, ranging from coarse-grained to fine-grained encodings, namely: LOBE, FRONTAL, GENERAL, and HEMISPHERE (see Appendix \ref{appendix:bre} for design ideas). Coarse-grained region division can aggregate more intra-region information, while fine-grained division can capture more inter-region information. The four schemes mentioned above are shown in Fig.~\ref{fig:region}.

We assign a brain region tag to each electrode, then map the tags into an embedding space using a learnable projection function, and simply add the embeddings to the node features. For node $i$, the encoding is defined as:
\begin{align}
\bm{r}_i &= \text{Embedding}(\text{Tag}({i})), \ \bm{r}_i \in \R^{d}, 
\end{align}
where $\text{Tag}$ identifies the brain region associated with node $i$, and nodes within the same brain region share a common embedding vector.

\subsubsection{Structure Encoding}
EEG channels are distributed in the 3D space, and the functional connectivity between them remains imprecisely defined. Therefore, inspired by \cite{graphormer3D}, we model the relationships between EEG channels as a fully connected directed graph to avoid making specific assumptions about functional connectivity.
Firstly, let $\phi(i,j)$ represent the Euclidean distance between node $i$ and node $j$, and encode $\phi(i,j)$ using a set of Gaussian basis functions \cite{graphormer3D,spinconv}. Let $\bm{b}_k \in \R^{n \times n}$ denote one of the Gaussian basis functions. The element $(i,j)$ of this function can be expressed as:
\begin{equation}
\bm{b}_k(i,j) = \mathcal{G}_k \rbr{\alpha_{ij} \phi(i, j) + \beta_{ij} - \mu_k, \sigma_k},
\end{equation}
where $\mu_k$ (mean), $\sigma_k$ (standard deviation), $\alpha_{ij}$, $\beta_{ij}$ are learnable scalars, and $i$ and $j$ denote the index of the source and target node, respectively.
By utilizing multiple Gaussian basis functions, we project the distance $\phi$ into various distributions.
A distance value close to the mean $\mu_k$ of the $k$-th basis function receives higher attention, while the standard deviation $\sigma_k$ controls the attention's concentration. This configuration provides flexibility and expressiveness to the encoding.
The outputs of these functions are concatenated, forming the structure encoding:
\begin{equation}
\bm{B} = \mathbin\Vert_{k=1} ^ K \bm{b}_k,\ \ \bm{B} \in \R^{n \times n \times K},
\end{equation}
where $\mathbin\Vert$ denotes the concatenation operator. 
Secondly, the structure encoding is summed along the second dimension, followed by an edge projection through a linear layer to derive the CE (see~\eqref{eqn:cen}) that reflects each node's relative importance. Nodes with higher cumulative weights are considered to have a greater level of involvement within the network.
The encoding process for a node in a sample of the input data is as follows:
\begin{align}
\label{eqn:cen}
\bm{c}_i &= \bm{e}_i \bm{W}_{\mathcal{E}}, \ \bm{e}_{i,k} = \sum_{j=1}^n \bm{B}_{i,j,k},\\
\bm{h}_i^{0} &= \tilde{\bm{x}}_i \bm{W}_{\mathcal{X}} + \bm{c}_i + \bm{r}_i,\\
\bm{B}' &= \text{Projection}(\bm{B}), 
\end{align}
where $i$ denotes the node index, and $k$ refers to the index of the basis function.  The matrix $\bm{W}_{\mathcal{X}} \in \R^{Tf \times d}$ represents a linear projection, while $\bm{W}_{\mathcal{E}} \in \R^{K \times d}$ is the edge projection.
$\text{Projection}: \mathbb{R}^{n \times n \times K} \mapsto \mathbb{R}^{n \times n \times M}$ is a nonlinear transformation for structure encoding to obtain the GSE, where $M$ is the number of attention heads. Finally, We incorporate this encoding as a bias term into the softmax attention (see~\eqref{eqn:bias}).

Unlike previous symmetric adjacency matrices \cite{EmoGT, rgnn}, our directed spatial encoding matrix allows the model to learn distinct correlations for $\rbr{i, j}$ and $\rbr{j, i}$.
Letting $l$ denote the model depth, $s$ denote the sample index and $m$ denote the index of multi-head attention, the multi-head attention module (MHA) can be represented as:\par
\begin{footnotesize}
\begin{gather}
\bm{Q}^{l, m}_s = \bm{H}^{l-1}_s \bm{W}_{\mathcal{Q}}^{l, m}, \bm{K}^{l, m}_s = \bm{H}^{l-1}_s \bm{W}_{\mathcal{K}}^{l, m}, \bm{V}^{l, m}_s=\bm{H}^{l-1}_s \bm{W}_{\mathcal{V}}^{l, m},\\
\label{eqn:bias}
\bm{Z}^{l, m}_s = \text{Softmax} \rbr{ \frac{\bm{Q}^{l, m}_s {\bm{K}^{l, m}_s}^\top}{\sqrt{d_h^{l,m}}} + \bm{B}'^{m}} \bm{V}^{l, m}_s,\\
\text{MHA}^{l}(\bm{H}^{l-1}_s) = \rbr{\mathbin\Vert_{m=1} ^ M \bm{Z}^{l, m}_s} \bm{W}_{\mathcal{O}}^{l},
\end{gather}
\end{footnotesize}%
where the projections $\bm{W}_{\mathcal{O}}^{l}$, $\bm{W}_{\mathcal{Q}}^{l, m}$, $\bm{W}_{\mathcal{K}}^{l, m}$ and $\bm{W}_{\mathcal{V}}^{l, m}$ are learnable model parameters. The scalar $d_h^{l,m}$ is the second dimension of $\bm{W}_{\mathcal{K}}^{l, m}$. 
The GSE bias term $\bm{B}'$ is incorporated into the softmax operation, enabling the model to adaptively adjust inter-channel relationships. For example, if $\bm{B}'(i, j)$ is a decreasing function, closer nodes will receive higher attention.

The proposed method integrates features from temporal, frequency, and spatial domains, expanding the model's capacity to represent complex EEG signals. By combining GSE with softmax attention, the model captures semantic correlations across nodes from multiple perspectives.

\subsection{Implementation Details of MVGT}
\label{sec:implementation}
In this section, we describe the overall architecture of the model, as illustrated in Fig.~\ref{fig:MVGT}. For better optimization, we first apply GraphNorm \cite{graphnorm} to normalize the input features to a range between 0 and 1.
The encodings could be characterized as below:
\begin{align}
\tilde{\bm{X}}_{s}' &= \text{GraphNorm}(\tilde{\bm{X}}_{s}),  \\
\bm{H}^{0}_s &= \text{Encoding}(\tilde{\bm{X}}_{s}').
\end{align}

We employ a Pre-LN structure, applying layer normalization (LN) before the MHA and the FFN.
This design choice is supported by prior research \cite{xiong2020on}, which demonstrates that Pre-LN structures produce more stable gradients, facilitating faster and more reliable convergence.
To further prevent overfitting, dropout is employed during training. The process is expressed mathematically as:
\begin{align}
\bm{H}'^{l}_s &= \text{MHA}^l(\text{LN}(\bm{H}^{l-1}_s)) + \bm{H}^{l-1}_s,  \\
\bm{H}^{l}_s &= \text{FFN}^l(\text{LN}({\bm{H}'^{l}_s})) + \bm{H}'^{l}_s.
\end{align}

Inspired by \cite{graphormer3D,AlphaFold}, we apply iterative refinement by recursively feeding the outputs of the model back into the same modules, denoted as ``recycling'' in Fig.~\ref{fig:MVGT}. The iterative approach refines the model's ability to discriminate encoded information and understand emotional patterns, thereby helping the model capture more effective details.

\section{Experiments}
\subsection{Datasets and Settings}
\label{sec:setting}
For our experiments, we select the SEED \cite{zheng2015investigating} and SEED-IV \cite{seediv} datasets to evaluate the effectiveness of our model. These datasets consist of EEG signals recorded from subjects watching emotion-eliciting videos.
Following the settings of previous studies \cite{zheng2015investigating,BiHDM,DGCNN,rgnn,MD-AGCN,EmoGT,seediv,li2022multi}, we use pre-computed differential entropy (DE) features for the recognition task.
Time segments are extracted using an overlapping sliding window of size $T$, consistent with \cite{MD-AGCN}, with $T$ set to 5.
For the SEED dataset, we use the first 9 trials of each subject as the training set and the last 6 trials as the test set, as done in previous research. The DE features are computed using five frequency bands extracted from 1s nonoverlapping windows. The model performance is evaluated based on the average accuracy and standard deviation across all subjects over two sessions of EEG data. 
Similarly, for the SEED-IV dataset, we allocate the first 16 trials to training and the last 8 trials to testing. The DE features for SEED-IV are calculated using 4s windows. The performance of our model is assessed using data from all three sessions. Refer to Appendix \ref{appendix:de} for DE feature extraction.

During experiments, the hidden dimension is set to 64 and the number of Gaussian basis functions is 32. The number of MHA layers is 4 and the number of attention heads is 2.
We employ iterative refinement by recursively feeding the output back into the model for three times. We set the batch size to 32 and the learning rate within the range of 3e-5 to 3e-3. Cross-entropy is used as the loss function, and AdamW \cite{adamw} is employed as the optimizer with a weight decay rate of 0.1.
\begin{table}[htb]
   \centering
   \caption{The classification accuracies (mean/std) on SEED and SEED-IV. MVGT-L, MVGT-G, MVGT-H, MVGT-F: MVGT using LOBE, GENERAL, HEMISPHERE and FRONTAL scheme. The best results are highlighted in bold, and the second-best results are underlined.}
   % \scalebox{0.80}{
   \begin{tabular}{c|c|c}
         \toprule
         \textbf{Model}&SEED&SEED-IV\\ \hline
         DGCNN \cite{DGCNN}&90.40/08.49&69.88/16.29\\
         BiDANN \cite{BiDANN}&92.38/07.04&70.29/12.63\\
         BiHDM \cite{BiHDM}&93.12/06.06&74.35/14.09\\
         R2G-STNN \cite{R2G-STNN}&93.34/05.96&-\\
         RGNN \cite{rgnn}&94.24/05.95&79.37/10.54\\
         MD-AGCN \cite{MD-AGCN}&94.81/04.52&87.63/05.77\\
         EmoGT \cite{EmoGT}&95.02/05.99&91.20/09.60\\ 
         MV-SSTMA \cite{li2022multi}&95.32/3.05&92.82/5.03\\ \hline
         MVGT-L&\underline{96.01}/04.85&\underline{93.20}/07.79\\
         MVGT-G&95.82/04.43&\textbf{94.03}/07.77\\
         MVGT-H&95.79/04.90&92.86/08.20\\
         MVGT-F&\textbf{96.55}/04.18&92.92/07.95\\
         \bottomrule
   \end{tabular}
   % }
   \label{tab:res}
\end{table}

\subsection{Results Analysis}
We compare the classification results on the SEED and SEED-IV datasets with recent state-of-the-art models, as shown in Table \ref{tab:res}. Under identical experimental conditions, our proposed model demonstrated significantly better results than the baselines.
On the SEED dataset, the FRONTAL scheme achieved the highest classification accuracy of 96.55\%, representing a 1.23\% improvement over the best baseline model. 
Similarly, on the SEED-IV dataset, the GENERAL scheme reached an accuracy of 94.03\%, exceeding the best baseline by 1.21\%.
The differences in accuracy across region division schemes likely result from their data-dependent nature.
This suggests that designing data-specific division schemes can enable the model to better utilize data characteristics, thereby enhancing recognition performance.
Conversely, improper schemes may result in poor performance.

\begin{figure}[hb]
   \centering
   \includegraphics[scale=0.15]{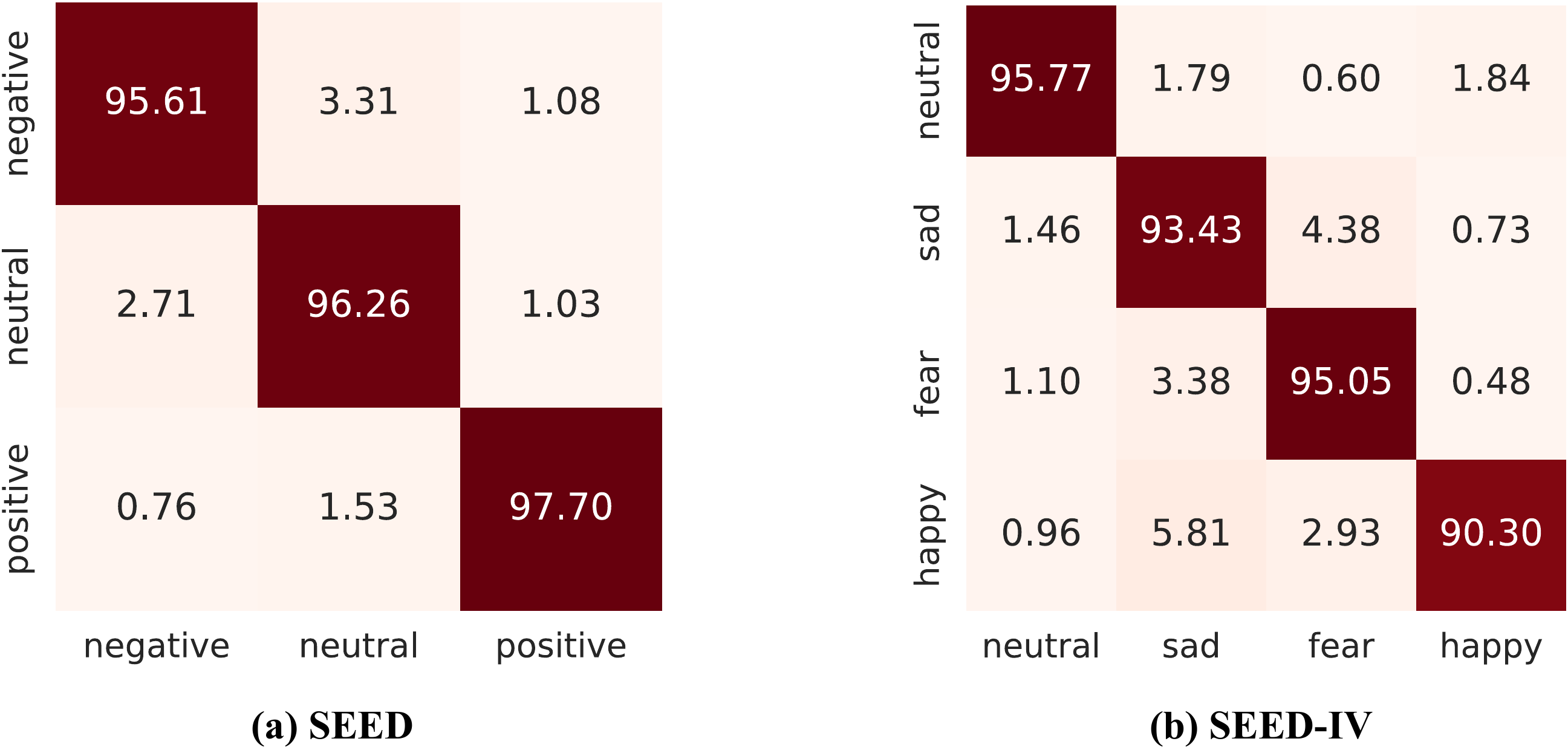}
   \caption{Confusion matrices of MVGT. (a) Confusion matrix of MVGT-F on SEED. (b) Confusion matrix of MVGT-G on SEED-IV. Each row of the matrix represents the true labels while each column serves as the predicted labels.}
   \label{fig:cf}
\end{figure}

Fig.~\ref{fig:cf} shows the confusion matrices for MVGT-F on SEED and MVGT-G on SEED-IV.
\begin{table}[htb]
   \centering
   \caption{Ablation study for the classification accuracies (mean/std) of MVGT-F on SEED and MVGT-G on SEED-IV. Symbol ``\yes'' indicates the component is employed.}
   \scalebox{0.85}{
         \begin{tabular}{cccccc}
         \toprule
         Centrality&BRE&GSE (bias \eqref{eqn:bias})&Inverted&SEED&SEED-IV\\ \hline
         -&-&-&-&92.49/07.58&87.40/11.36\\ \hline
         -&-&-&\yes&93.29/06.93&88.11/10.30\\ \hline
         -&\yes&-&\yes&94.02/06.30&88.65/10.39\\ \hline
         \yes&-&-&\yes&94.01/05.96&89.25/09.49\\ \hline
         \yes&\yes&-&\yes&94.17/05.33&89.58/09.25\\ \hline
         -&-&\yes&\yes&93.79/07.15&89.49/10.40\\ \hline
         -&\yes&\yes&\yes&95.10/05.01&91.46/09.75\\ \hline
         \yes&-&\yes&\yes&95.05/05.09&92.82/07.95\\ \hline
         \yes&\yes&\yes&\yes&\textbf{96.55/04.18}&\textbf{94.03/07.77}\\
         \bottomrule
         \end{tabular}
   }
   \label{tab:ablation}
\end{table}
For SEED, the model achieves 97.70\% accuracy for positive emotions but performs less effectively on negative emotions (95.61\%). Misclassification rates are low, with 0.76\% of positive samples misclassified as negative and 1.08\% of negative samples as positive, demonstrating effectiveness in distinguishing valence changes. For SEED-IV, the model achieves its best performance on neutral emotions (95.77\%) but slightly lower accuracy on happy emotions (90.30\%), likely due to the GENERAL scheme's sensitivity to balanced emotions.

\subsection{Ablation Study}
To validate the effectiveness of our proposed method, we conduct ablation experiments on MVGT-F (based on SEED) and MVGT-G (based on SEED-IV).
These experiments incrementally remove four key components: centrality encoding (CE), brain region encoding (BRE), geometric structure encoding (GSE), and the inverted temporal embedding (Inverted). GSE is incorporated into the attention mechanism as a bias (as detailed in \eqref{eqn:bias}).
The term ``Inverted'' refers to embedding continuous time segments as tokens (described in Section \ref{sec:time}), whereas the default approach treats multi-channel information from a single time point as tokens.

Applying the ``Inverted'' method to the default pure Transformer improves model accuracy by 0.8\% on SEED and 0.71\% on SEED-IV, highlighting the benefit of encoding continuous temporal information. Among the spatial encoding components, GSE has the greatest impact. Its inclusion in the attention mechanism increases accuracy by an average of 1.25\% on SEED and 3.05\% on SEED-IV.
In comparison, incorporating BRE improves accuracy by 0.93\% (SEED) and 1.01\% (SEED-IV), while CE contributes average accuracy gains of 0.90\% (SEED) and 1.99\% (SEED-IV).
These results demonstrate the effectiveness of the temporal embedding method and spatial encoding components in enhancing the model's ability to classify emotion labels. Moreover, integrating all four components maximizes the model's overall performance, underscoring their complementary roles in capturing complex EEG patterns.
\begin{figure*}[htb]
   \centering
   \includegraphics[scale=0.39]{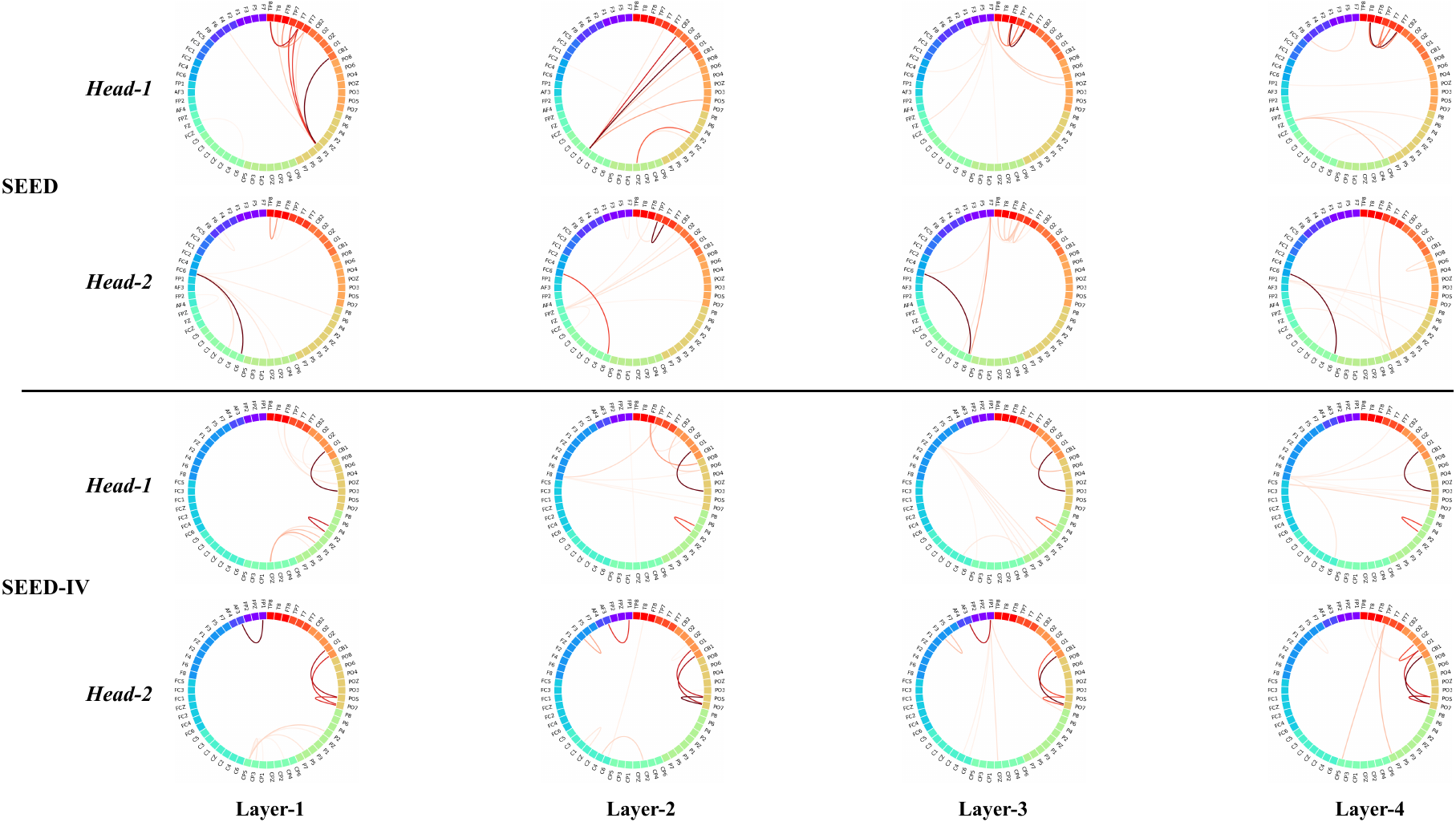}
   \caption{The learned inter-channel relationships from the SEED by the MVGT-F and from the SEED-IV by the MVGT-G are illustrated. The figures show the results of the last iteration in the iterative refinement, highlighting the top 10 channel pairs with the highest weights after softmax (darker colors indicate higher weights). Channels of the same brain region are represented in the same color. Rows correspond to attention heads, while columns represent the layers of the MHA mechanism.}
   \label{fig:connectivity}
\end{figure*}

\subsection{Visualization of Inter-channel Relations}
To better illustrate the inter-channel correlations, we visualize the channel relationships captured by MVGT-F on the SEED dataset and MVGT-G on the SEED-IV dataset. From the final MHA layer of the last iterative refinement, we select the top 10 strongest connections. Fig.~\ref{fig:connectivity} displays the average weights across all subjects.

On the SEED dataset, the parameters suggest that emotional patterns involve activity across multiple brain regions. The parameters from head-1 in MVGT-F suggest that brain activity is primarily concentrated in the temporal and frontal regions on the lateral side of the brain, which aligns with previous research \cite{zheng2015investigating,rgnn,EmoGT}. Notably, the channel connections (TP8, T7), (FT8, T7), and (T8, FT7) reveal a strong correlation between the left and right hemispheres.
The connection (FC6, C6) in head-2 shows the strongest activity, further indicating that local information in the central region may play a key role in emotion recognition.

For the SEED-IV dataset, the strongest inter-channel connections occur in the frontal, parietal, and occipital regions, consistent with existing findings \cite{rgnn}. In MVGT-G, the main channel connections are (O1, PO3), (P4, P2), (CB1, PO7), and (PO5, PO7). Additionally, the connection between AF3 and FP1 provides critical insights into emotion processing, in line with \cite{rgnn,EmoGT}.

Our model incorporates both intra-regional and inter-regional brain dynamics, highlighting the critical role of distributed network interactions in modeling emotional states, as discussed by \cite{barrett2013large}.

\section{Conclusions}
This paper presents a multi-view graph transformer (MVGT) based on spatial relations for EEG-based emotion recognition. Our model systematically leverages the frequency, temporal, and spatial properties of EEG data. In the frequency domain, we utilize differential entropy of EEG signals as the basis for emotion recognition. In the temporal domain, we extend the temporal receptive field to capture temporal dynamics from continuous time segments. Additionally, we integrate three spatial encodings into the model to enhance its expressive power and adaptability to spatial structures. Extensive experiments on public emotion recognition datasets demonstrate that our proposed model outperforms other competitive baselines. 
Furthermore, an analysis of inter-channel correlations indicates that emotional brain activity emerges from the coordinated interaction of multiple brain regions rather than isolated areas. 

\section*{Acknowledgement}
This preprint has not undergone peer review (when applicable) or any post-submission improvements or corrections.
The Version of Record of this contribution is published in LNCS 16310, and is available online at \url{https://doi.org/10.1007/978-981-95-4378-6_1}

\section{Appendix}
\subsection{DE Feature Extraction}
\label{appendix:de}
Differential Entropy (DE) extends Shannon entropy to its continuous form and is used to measure the complexity of a continuous random variable. It is typically calculated based on precomputed frequency band information, which can be obtained using the Short-Time Fourier Transform (STFT). Let the original EEG signal be denoted as $X \in \mathbb{R}^{C \times N}$, which is divided into $W$ time windows, each containing: $L={\frac{N}{W}} $ sampling points. The data for the $w$-th window is:
\begin{equation}
\mathbf{X}_{w}=\mathbf{X}[:, w\cdot L:(w+1)\cdot L].
\end{equation}
A Hanning window function is applied to the data in each window, defined as:
\begin{equation}
h[n]=0.5\left(1-\cos\left({\frac{2\pi n}{L-1}}\right)\right),\quad n=0,1,\cdot\cdot\cdot,L-1
\end{equation}
the signal after applying the window is:
\begin{equation}
{\bf X}_{w}^{(h)} = {\bf X}_{w}\cdot h.
\end{equation}
Next, the Hanning-windowed signal undergoes an $N_{\mathrm{FFT}}$-point Fourier Transform to obtain the frequency-domain signal:
\begin{equation}
\mathbf{F}_{w}=\mathbf{F}\mathbf{F}\mathbf{T}(\mathbf{X}_{w}^{(h)},N_{\mathrm{FFT}}),
\end{equation}
the magnitude of the positive frequencies of the spectrum is taken as:
\begin{equation}
\mathbf{E}_{w}=|\mathbf{F}_{w}[:,0:{\frac{N_{\mathrm{FFT}}}{2}}]|,
\end{equation}
\begin{figure*}[htb]
    \centering
    \includegraphics[scale=0.09]{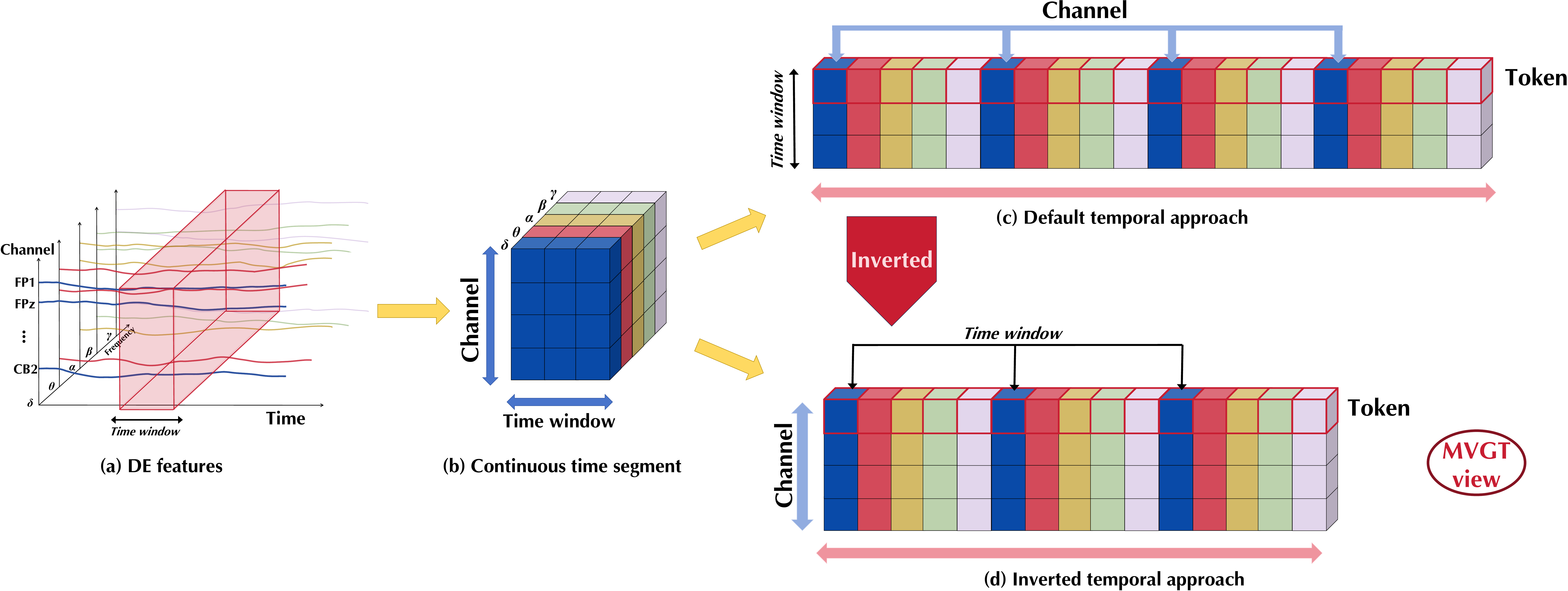}
    \caption{The temporal approach is illustrated. (a) represents the DE data, (b) represents the time segments obtained by a sliding window, (c) shows the default method of treating multi-channel data at a single time point as a token, while (d) illustrates the ``Inverted'' method, where the entire continuous time segment is treated as a token.}
    \label{fig:time}
\end{figure*}
The average power spectral density (PSD) for a frequency band $b$ (from $f_{\mathrm{low}}$ to $f_{\mathrm{high}}$) is computed. The start and end indices for the band are:
\begin{equation}
k_{\mathrm{start}}=\left \lfloor {\frac{f_{\mathrm{low}}}{f_{s}}}\cdot N_{\mathrm{FFT}}\right \rfloor,\quad k_{\mathrm{end}}=\left\lfloor {\frac{f_{\mathrm{high}}}{f_{s}}}\cdot N_{\mathrm{FFT}}\right \rfloor, 
\end{equation}
where $f_s$ is the sampling frequency. The average PSD is given by:
\begin{equation}
\mathrm{PSD}_{w,b}={\frac{1}{k_{\mathrm{end}}-k_{\mathrm {start}}+1}}\sum_{k=k_{\mathrm{start}}}^{k_{\mathrm{end}}}\mathbf{E}_{w}[:k]^{2}.
\end{equation}
Finally, the DE feature is computed as
\begin{equation}
{\mathrm{DE}}_{w,b}=\log_{2}(100\cdot{\mathrm{PSD}}_{w,b}).
\end{equation}
Typically, DE features are computed for the frequency bands $\delta$ (1-4 Hz), $\theta$ (4-8 Hz), $\alpha$ (8-14 Hz), $\beta$ (14-31 Hz), and $\gamma$ (31-50 Hz). A single sample of the resulting DE features is represented as a $C \times 5$ vector.

\subsection{Temporal Embedding}
\label{appendix:time}
To capture the temporal dynamics of EEG signals, we slice the signals along the time axis using time windows (as shown in Fig.~\ref{fig:time}). In this figure, (a) represents the original DE data, (b) represents the time segments obtained by slicing using a sliding window, (c) shows the default temporal embedding of treating multi-channel data at a single time point as a token, while (d) illustrates the ``Inverted'' method, where the entire continuous time segment is treated as a token, allowing a feedforward neural network to capture the continuous temporal dynamics.

\subsection{Datasets and Baselines}
\label{appendix:dataset}
\subsubsection{Datasets}
\textbf{SEED} dataset comprises data from 15 subjects who participated in three sessions, each separated by at least one week. Each session consists of 15 trials capturing emotional labels, with the emotion labels being positive, negative, and neutral.

\textbf{SEED-IV} dataset is constituted by EEG signals from 15 subjects across three separate sessions conducted at different times, using the same device as the SEED dataset. This dataset encompasses four emotion labels: neutral, sad, fear, and happy. In each session, each subject underwent 24 trials.

\subsubsection{Baseline Models}
DGCNN \cite{DGCNN}: A dynamic graph neural network method based on Chebyshev polynomials dynamically learns inter-channel relations in emotion recognition.

BiDANN \cite{BiDANN}: The bi-hemispheres domain adversarial neural network uses recurrent neural networks to capture temporal features and extracts hemisphere information by modeling the left and right hemispheres.

BiHDM \cite{BiHDM}: This model employs a pairwise subnetwork to capture the discrepancy between the left and right hemispheres of the brain.

R2G-STNN \cite{R2G-STNN}: A model that captures spatial-temporal features from local to global scales for emotion classification.

RGNN \cite{rgnn}: A regularized GNN that learns topological relationships between channels.

MD-AGCN \cite{MD-AGCN}: An adaptive GNN that comprehensively considers temporal domain, frequency domain, and brain functional connectivity.

EmoGT \cite{EmoGT}: An elastic graph Transformer network that integrates temporal and spatial information of EEG signals.

MV-SSTMA \cite{li2022multi}: A multi-view masked autoencoder combining CNN and Transformer for emotion recognition.

\subsection{Brain Region Encoding}
\label{appendix:bre}
The partitioning schemes are based on the following considerations:
\begin{itemize}
\item[$\bullet$] We divide the regions based on the anatomical structure of the brain and implement LOBE scheme.
\item[$\bullet$] To further investigate the expressive power of brain region encoding, we conduct a detailed division of brain lobes according to the 10-20 system based on electrode positions, employing the GENERAL scheme.
\item[$\bullet$] Asymmetric EEG activity in the frontal lobe can be utilized for discriminating valence changes \cite{schmidt2001frontal}. The left frontal lobe exhibits a stronger correlation with joy and happiness, while the right frontal lobe is more strongly correlated with fear and sadness. Thus we further divide the frontal lobe region into two symmetrical regions to obtain the FRONTAL scheme.
\item[$\bullet$] According to the symmetry of brain structure \cite{grabner2012oscillatory}, we make a finer division of the brain lobe regions, defining the HEMISPHERE scheme.
\end{itemize}

% \subsection{Complexity}
% The main computational bottleneck of the MVGT model lies in the Multi-Head Attention (MHA) module, with a time complexity of $\bm {O(CLMV^2d)}$, where $V$ is the number of nodes, $M$ is the number of attention heads, $L$ is the number of MHA layers, $C$ is the number of iterative refinements, and $d$ is the feature dimension of a head. The overall computational complexity is $\bm {O(V^2d)}$, which is on the same scale as \cite{EmoGT,li2022multi}. For graph convolution-based methods, the complexity depends on the number of edges: the complexity for sparse graphs is $\bm {O(Ed)}$, while for dense graphs, the complexity is $\bm {O(V^2d)}$.

\subsection{Limitation}
The brain region division we designed is based on neuroscience research on emotional processes, and the model's performance depends on the encoding design—improper encoding may introduce noise.
In addition, since the brain region encoding of MVGT is based on conclusions from emotional research, it may not necessarily be applicable to EEG datasets from other cognitive tasks.

\bibliographystyle{splncs04}
\bibliography{reference}
\end{document}